\documentclass[runningheads]{llncs}
\pdfoutput=1

\usepackage{graphicx}
\usepackage{amsmath,amssymb} 
\usepackage{color}
\usepackage[width=122mm,left=12mm,paperwidth=146mm,height=193mm,top=12mm,paperheight=217mm]{geometry}
\usepackage[group-separator={,}]{siunitx}

\usepackage{acronym}
\newacro{AP}[AP]{average precision}
\newacro{artwork}[artwork]{non-natural images}
\newacro{BoVW}[BoVWs]{Bag of Visual Words}
\newacro{CNN}[CNN]{convolutional neural network}
\newacro{DPM}[DPM]{Deformable Part-based Model}
\newacro{FastRCNN}[Fast R-CNN]{the ``Fast Region-based Convolutional Network'' method}
\newacro{HOG}[HOG]{Histograms of Oriented Gradients}
\newacro{IOU}[IoU]{intersection over union}
\newacro{photo}[photo]{natural image}
\newacro{RCNN}[R-CNN]{``regions with CNN features''}
\newacro{ROI}[ROI]{region of interest}
\newacro{RELU}[ReLU]{rectified linear unit}
\newacro{SGD}[SGD]{stochastic gradient descent}
\newacro{SPP}{spatial pyramid pooling}
\newacro{SVM}[SVM]{support vector machine}
\newacro{SSVM}[SSVM]{structured support vector machine}
\newacro{TSNE}[t-SNE]{t-Distributed Stochastic Neighbour Embedding}
\newacro{VOC}[VOC]{Visual Object Classes}
\newacro{YOLO}[YOLO]{``You Only Look Once''}

\begin{document}
\pagestyle{headings}
\mainmatter
\def\ECCV16SubNumber{18}  

\title{Detecting People in Artwork with CNNs}

\titlerunning{Detecting People in Artwork with CNNs}
\authorrunning{Nicholas Westlake, Hongping Cai, Peter Hall}

\author{Nicholas Westlake\inst{1}, Hongping Cai\inst{2}, and Peter Hall\inst{1}}
\institute{Department of Computer Science, Unversity of Bath, Bath, UK\\
  \email{ \{n.westlake,p.m.hall\}@bath.ac.uk}\and
  Department of Computer Science, University of Bristol, Bristol, UK\\
  \email{ hongping.cai@bristol.ac.uk} }

\maketitle

\begin{abstract}
  CNNs have massively improved performance in object detection in photographs.
  However research into object detection in artwork remains limited.
  We show state-of-the-art performance on a challenging dataset, \emph{People-Art}, which contains people from photos, cartoons and 41 different artwork movements.
  We achieve this high performance by fine-tuning a CNN for this task, thus also demonstrating that training CNNs on photos results in overfitting for photos: only the first three or four layers transfer from photos to artwork.
  Although the CNN's performance is the highest yet, it remains less than 60\% AP, suggesting further work is needed for the cross-depiction problem.
\end{abstract}

\section{Introduction}
\begin{figure}[b]
    \includegraphics[height=2.3cm]{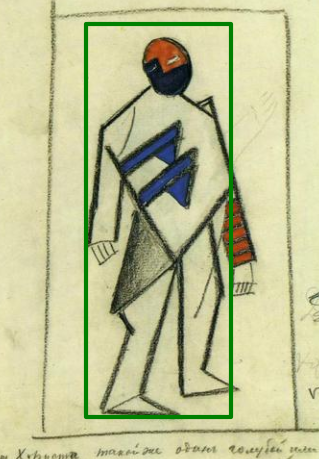} \hfill
    \includegraphics[height=2.3cm]{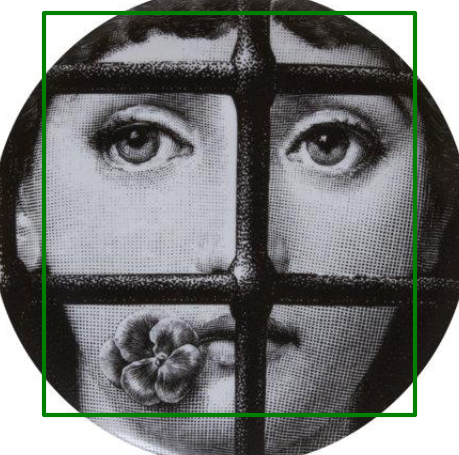} \hfill
    \includegraphics[height=2.3cm]{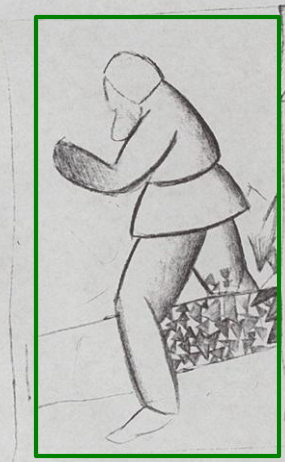} \hfill
    \includegraphics[height=2.3cm]{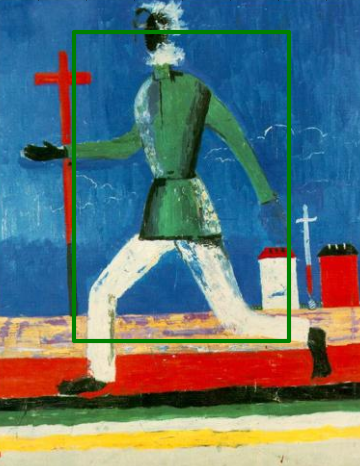} \hfill
    \includegraphics[height=2.3cm]{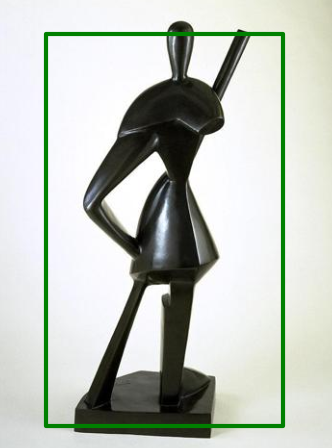} \hfill
    \includegraphics[height=2.3cm]{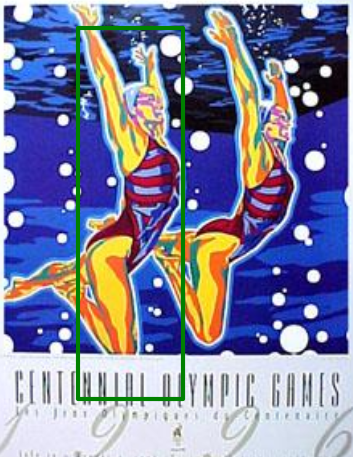}
\caption{Detecting people across different depictive styles a challenge: here we show some successful detections.}
\end{figure}

Object detection has improved significantly in recent years, especially as a result of the resurgence of \acp{CNN} and the increase in performance and memory of GPUs.
However, in spite of the successes in photo-based recognition and detection, research into recognition within styles of images other than \acp{photo} remains limited~\cite{hall2015cross}.
We refer to this as the \textit{cross-depiction problem}: detecting objects regardless of how they are depicted (photographed, painted, drawn, etc.).

We believe that cross-depiction recognition is an interesting and open problem.
It is interesting because it forces researchers to look beyond the surface appearance of object classes.
By analogy, just as a person retains their identity no matter what clothes they wear, so an object retains its class identity no matter how it is depicted:
a dog is a dog whether photographed, painted in oils, or drawn with a stick in the sand.

Cross-depiction is a practical problem too: an example is an image search.
The world contains images in all sorts of depictions.
Any recognition solution that does not generalise across these depictions is of limited power.
Yet most current computer vision methods tacitly assume a photographic input, either by design or training.
Any model premised on a single depictive style e.g.\ \acp{photo} will lack sufficient descriptive power for cross-depiction recognition.
Therefore, an image search using methods will limit its results to \acp{photo} and photo-like depictions.

In our paper, we talk about \acfp{photo} and \ac{artwork} as a linguistic convenience.
We would argue that this is a false dichotomy: the universe of all images includes images in all possible depictive styles, and there is no particular reason to privilege any one style.
Nevertheless, we acknowledge that the distribution of styles is not uniform: \acp{photo} may be more abundant and certainly are in computer vision datasets such as ImageNet~\cite{deng2009imagenet}.
This creates problems for generalisation: training a detector on \acp{photo} alone constrains it not only in terms its ability to handle denotational varieties, but projective and pose varieties too, as we discuss later.

We present a new dataset, \emph{People-Art}, which contains \acp{photo}, cartoons and images from 41 different \ac{artwork} movements.
Unlike the \emph{Photo-Art} dataset~\cite{wu2014learning}, which had 50 classes, this dataset has a single class: people.
We labelled people since we observe that people occur far more frequently across the wide spectrum of depictive styles than other classes, thus allowing a far greater variety.
Detecting people within this dataset is a challenging task because of the huge range of ways artists depict people: from Picasso's cubism to Disney's Sleeping Beauty.
The best performance on a pre-release of the dataset is 45\% \ac{AP}, from a \ac{CNN} that was neither trained nor fine-tuned for this task.
By fine-tuning a state-of-the-art \ac{CNN} for this task~\cite{girshick2015fast}, we achieved 58\% \ac{AP}, a substantial improvement.

As well as achieving state-of-art performance on our \textit{People-Art} dataset,
we make the following contributions, in order of strength:
\begin{enumerate}
\item We show that a simple tweak for \ac{FastRCNN}~\cite{girshick2015fast}, changing the criteria for negative training exemplars compared to default configuration, is key to higher performance on artwork.
\item We show the extent to which fine-tuning a \ac{CNN} on \ac{artwork} improves performance when detecting people in \ac{artwork} on our dataset (Section \ref{sec:peopleArtBenchmarks}) and the \textit{Picasso} dataset~\cite{ginosar2014detecting} (Section \ref{sec:PicassoBenchmarks}).
  We show that this alone is not a solution: the performance is still less than 60\% AP after fine tuning, suggesting the need for futher work.
\item Consistent with earlier work~\cite{yosinski2014transferable}, we show that the lower convolutional layers of a \ac{CNN} generalise to artwork: others benefit from fine-tuning (Section \ref{sec:ROISelection}).
\end{enumerate}
We begin by presenting related work and our \textit{People-Art} dataset.

\begin{figure}[p]
  \tiny
  \newcolumntype{C}{>{\centering\arraybackslash}p{0.175\textwidth}}
  \begin{tabular}{CCCCC}
    \includegraphics[width=0.175\textwidth]{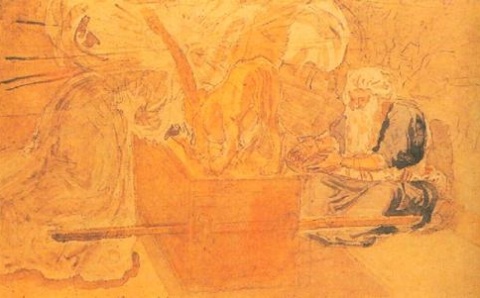} &
    \includegraphics[width=0.175\textwidth]{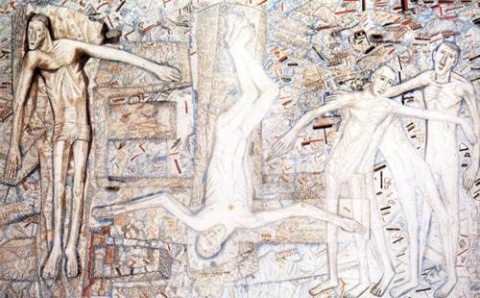} &
    \includegraphics[width=0.175\textwidth]{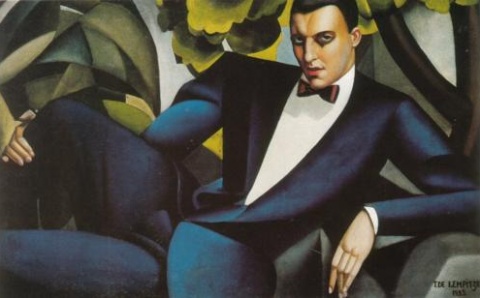} &
    \includegraphics[width=0.175\textwidth]{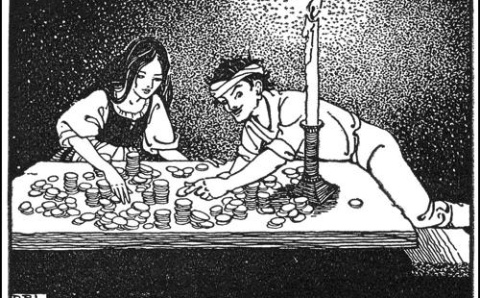} &
    \includegraphics[width=0.175\textwidth]{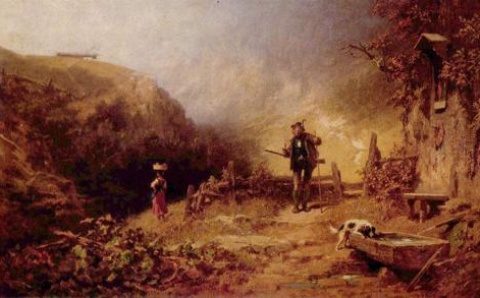}
    \\
    Academicism & AnalyticalRealism & Art Deco & Art Nouveau & Biedermeier
    \\[8pt]
    \includegraphics[width=0.175\textwidth]{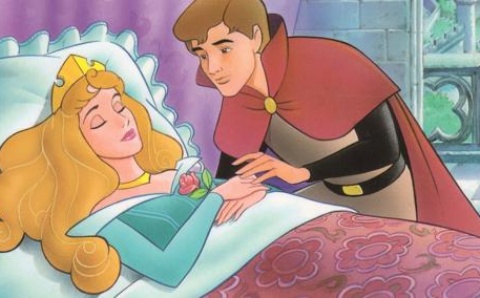} &
    \includegraphics[width=0.175\textwidth]{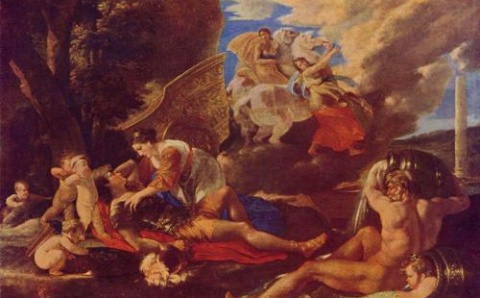} &
    \includegraphics[width=0.175\textwidth]{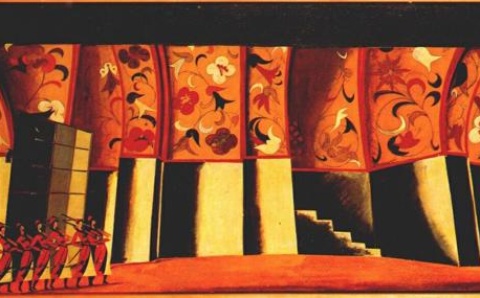} &
    \includegraphics[width=0.175\textwidth]{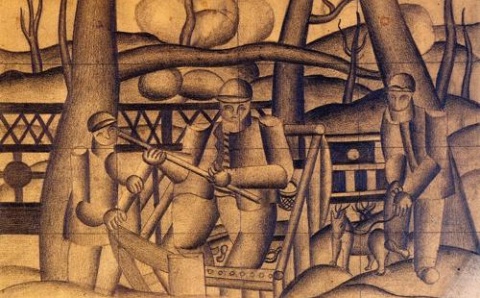} &
    \includegraphics[width=0.175\textwidth]{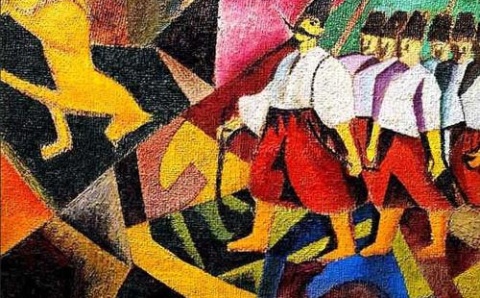}
    \\
    cartoon & classicism & constructivism & Cubism & Cubo-Futurism
    \\[8pt]
    \includegraphics[width=0.175\textwidth]{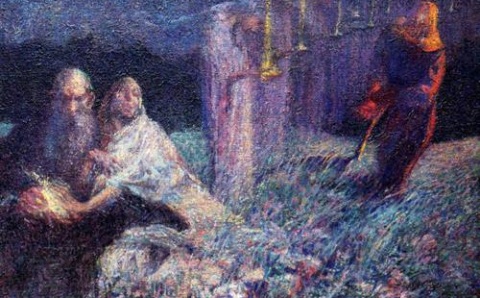} &
    \includegraphics[width=0.175\textwidth]{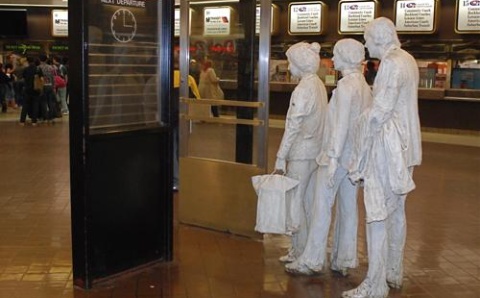} &
    \includegraphics[width=0.175\textwidth]{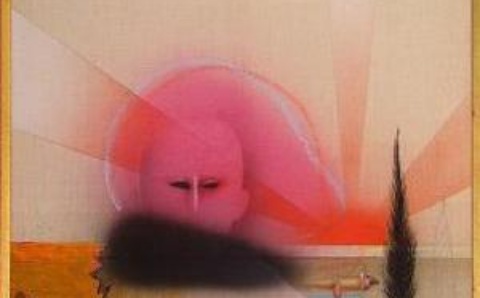} &
    \includegraphics[width=0.175\textwidth]{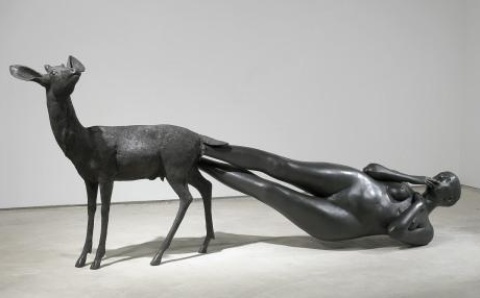} &
    \includegraphics[width=0.175\textwidth]{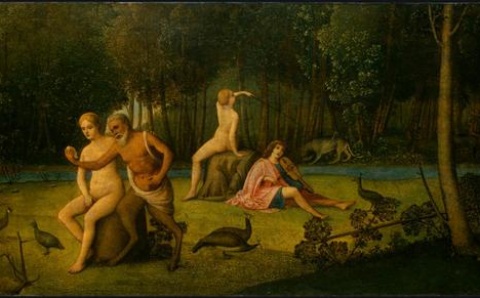} 
    \\
    Divisionism & Environmental Art & fantastic realism & feminist art & High Renaissance 
    \\[8pt]
    \includegraphics[width=0.175\textwidth]{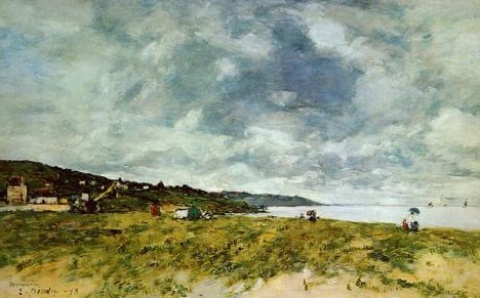} &
    \includegraphics[width=0.175\textwidth]{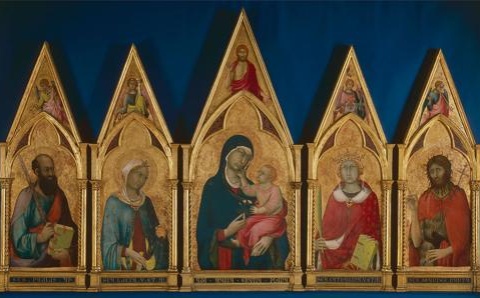} &
    \includegraphics[width=0.175\textwidth]{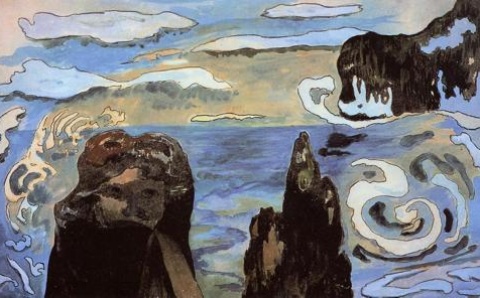} &
    \includegraphics[width=0.175\textwidth]{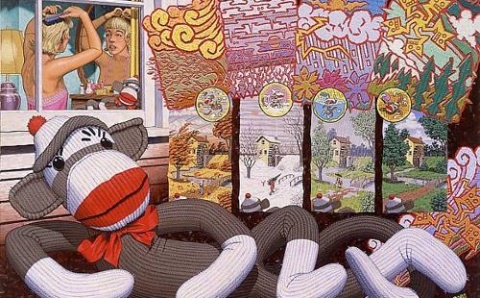} &
    \includegraphics[width=0.175\textwidth]{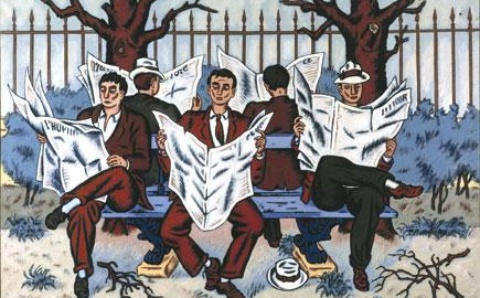} 
    \\
    impressionism & International Gothic & Japonism & lowbrow art & magic realism
    \\[8pt]
    \includegraphics[width=0.175\textwidth]{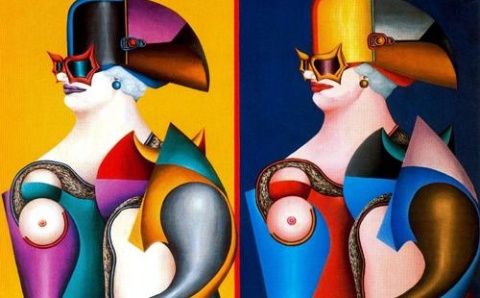} &
    \includegraphics[width=0.175\textwidth]{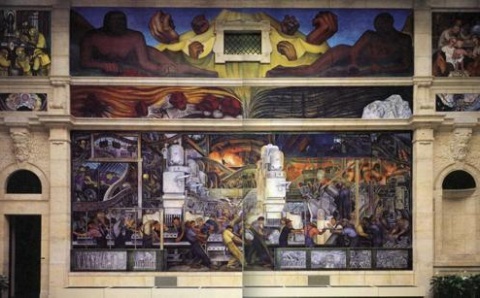} &
    \includegraphics[width=0.175\textwidth]{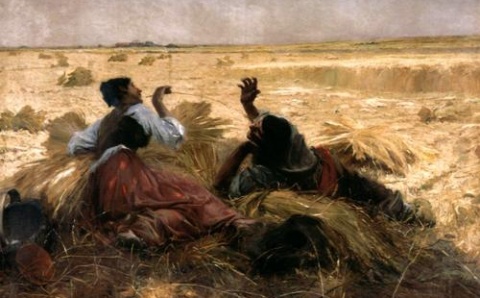} &
    \includegraphics[width=0.175\textwidth]{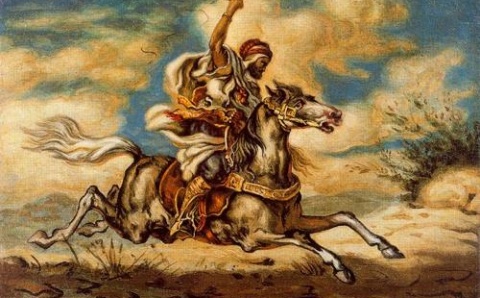} &
    \includegraphics[width=0.175\textwidth]{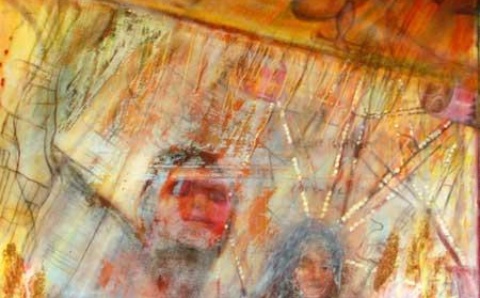} 
    \\
    Mechanistic Cubism & muralism & Naturalism & Neo-Baroque & neo-figurative art
    \\[8pt]
    \includegraphics[width=0.175\textwidth]{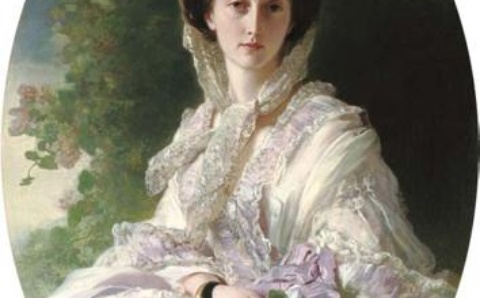} &
    \includegraphics[width=0.175\textwidth]{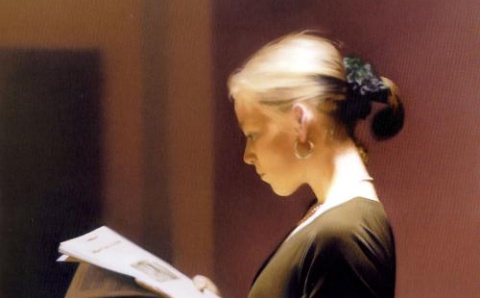} &
    \includegraphics[width=0.175\textwidth]{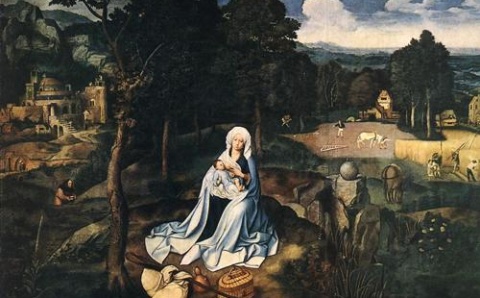} &
    \includegraphics[width=0.175\textwidth]{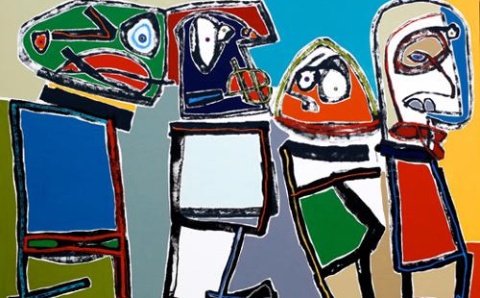} &
    \includegraphics[width=0.175\textwidth]{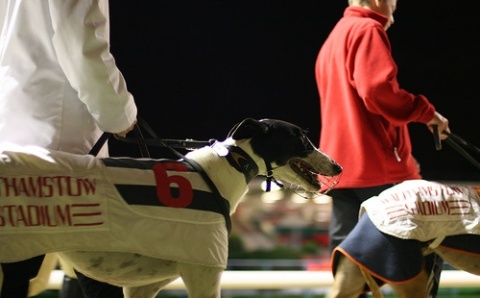} 
    \\
    neo-rococo & New European Painting & Northern Renaissance & outsider art & photo
    \\[8pt]
    \includegraphics[width=0.175\textwidth]{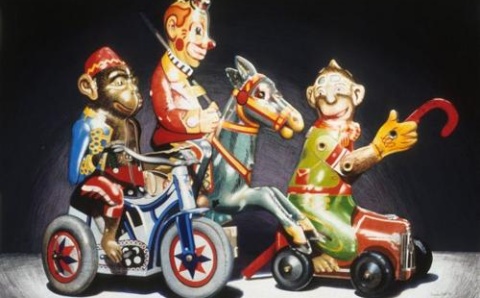} &
    \includegraphics[width=0.175\textwidth]{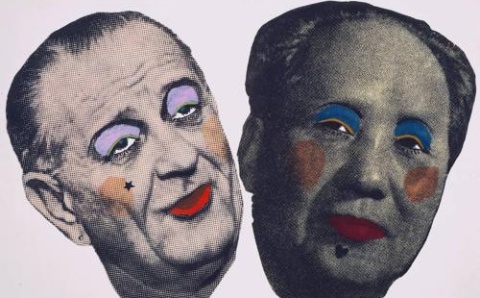} &
    \includegraphics[width=0.175\textwidth]{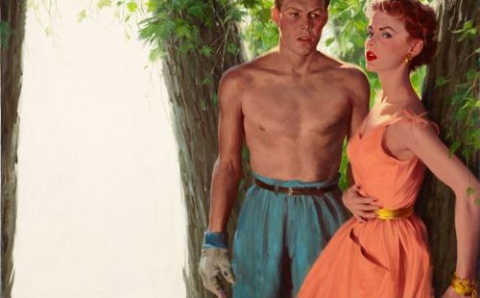} &
    \includegraphics[width=0.175\textwidth]{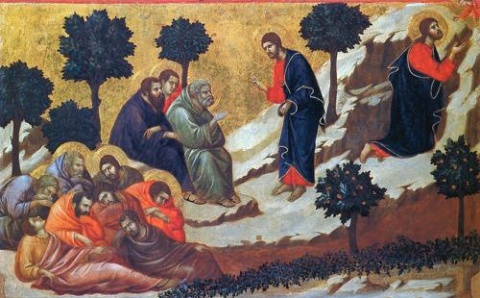} &
    \includegraphics[width=0.175\textwidth]{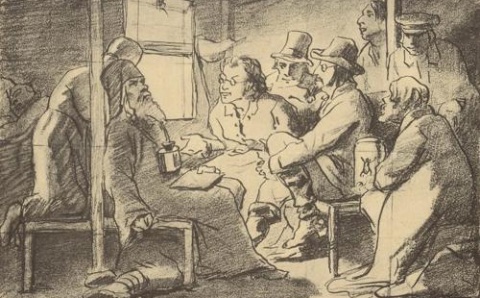} 
    \\
    photorealism & pop art & poster art realism & Proto-Renaissance & realism 
    \\[8pt]
    \includegraphics[width=0.175\textwidth]{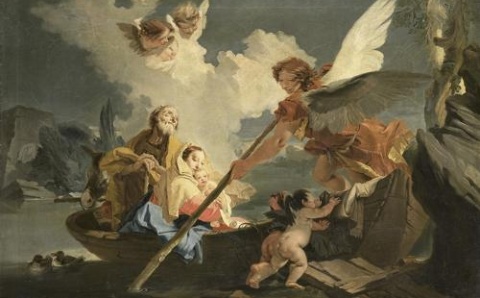} &
    \includegraphics[width=0.175\textwidth]{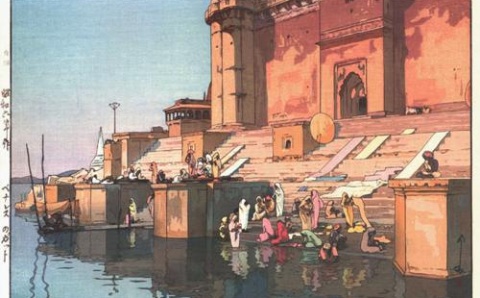} &
    \includegraphics[width=0.175\textwidth]{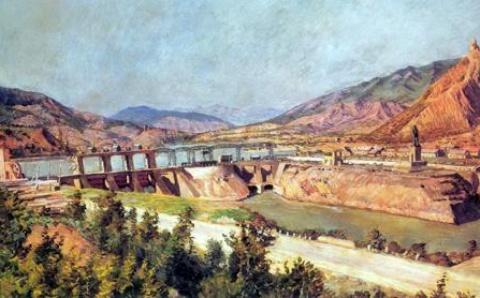} &
    \includegraphics[width=0.175\textwidth]{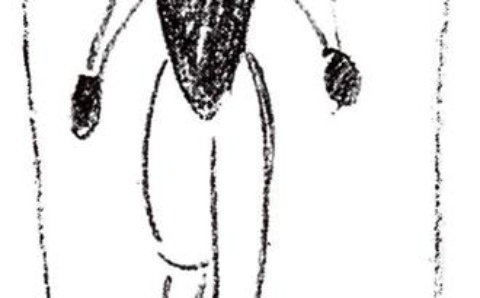} &
    \includegraphics[width=0.175\textwidth]{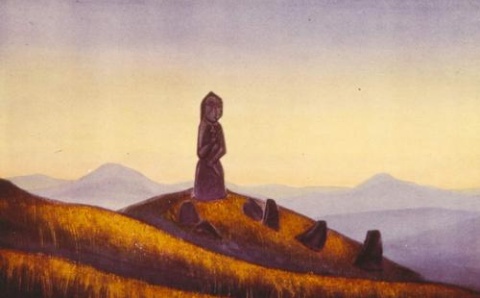} 
    \\
    rococo & shin-hanga & socialist realism & Suprematism & symbolism 
    \\[8pt]
    &
    \includegraphics[width=0.175\textwidth]{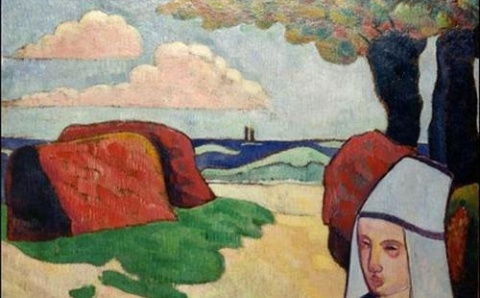} &
    \includegraphics[width=0.175\textwidth]{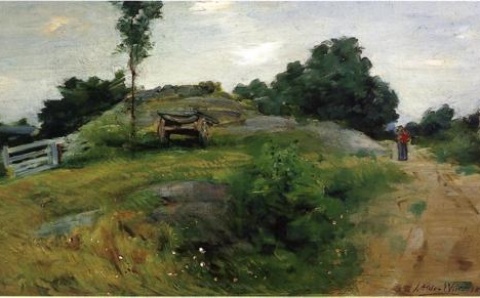} &
    \includegraphics[width=0.175\textwidth]{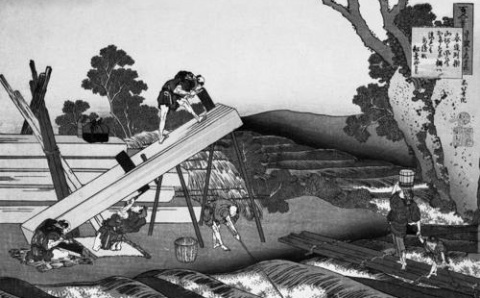} 
    \\
    & synthetism & Tonalism & ukiyo-e  
  \end{tabular}
  \normalsize
  \caption{Our \textit{People-Art} dataset contain  images from 43 different styles of depiction: here we show one example for depiction style.\label{fig:oneImagePerStyle}}
\end{figure}

\section{Related Work}
We use a state-of-the-art \ac{CNN} to improve performance on a cross-depiction dataset, thereby contributing towards cross-depiction object recognition.
We first explore related work on deep learning for object detection and localisation (largely in \acp{photo}), followed by previous work on the cross-depiction problem.

\subsection{Deep Learning for Object Detection and Localisation}
Deep learning has been around for a few decades~\cite{fukushima1980neocognitron,giebel1971feature,lecun1989backpropagation}.
After a period of limited use within computer vision, Krizhevsky et al.\ (2012)~\cite{krizhevsky2012imagenet} demonstrated a vast performance improvement for image classification over previous state-of-the-art methods, using a deep \ac{CNN}.
As a result, the use of \acp{CNN} surged within computer vision.

Early \ac{CNN} based approaches for object localisation~\cite{matan1992reading,nowlan1995convolutional,rowley1998neural,sermanet2013pedestrian} used the same sliding-window approach used by previous state-of-the-art detection systems~\cite{dalal2005histograms,felzenszwalb2010object}.
As \acp{CNN} became larger, and with an increased number of layers, this approach became intractable.
However, Sermanet et al. (2014)~\cite{sermanet2014overfeat} demonstrated that few windows are required, provided the \ac{CNN} is fully convolutional.
Furthermore, as the size of their receptive fields increased, \acp{CNN} either became or were trained to be less sensitive to precise location and scale the input.
As a result, obtaining a precise bounding box using sliding window and non-maximal suppression became difficult.
One early approach attempted to solve this issue by training a separate \ac{CNN} for precise localisation~\cite{vaillant1994original}.

Szegedy et al. (2013)~\cite{szegedy2013deep} modified the architecture of Krizhevsky et al.\ (2012)~\cite{krizhevsky2012imagenet} for localisation by replacing the final layer of the \ac{CNN} with a regression layer.
This layer produces a binary mask indicating whether a given pixel lies within the bounding box of an object.
Schulz and Behnke (2011)~\cite{schulz2011object} previously used a similar approach with a much smaller network for object segmentation.

Girshick et al. (2014)~\cite{girshick2014rich} introduced \ac{RCNN}, which surpassed previous approaches.
The authors used selective search~\cite{uijlings2013selective}, a hierarchical segmentation method, to generate region proposals: possible object locations within an image.
Next, a \ac{CNN} obtains features from each region and a \ac{SVM} classifies each region.
In addition, they used a regression model to improve the accuracy of the bounding box output by learning bounding box adjustments for each class-agnostic region proposal.
He et al. (2015)~\cite{he2015spatial} improved the run-time performance by introducing SPP-net, which uses a \ac{SPP}~\cite{grauman2005pyramid,lazebnik2006beyond} layer after the final convolutional layer.
The convolutional layers operate on the whole image, while the \ac{SPP} layer pools based on the region proposal to obtain a fixed length feature vector for the fully connected layers.

Girshick (2015)~\cite{girshick2015fast} later introduced \ac{FastRCNN} which improves upon \ac{RCNN} and SPP-net and allows the \ac{CNN} to output a location of the bounding box (relative to the region proposal) directly, along with class detection score, thus replacing the \ac{SVM}.
Furthermore, this work enables end-to-end training of the whole \ac{CNN} for both detection and bounding box regression.
We use this approach to achieve state-of-the-art performance on our \emph{People-Art} dataset and detail the method in Section \ref{sec:CNNarchitecture}.

To make \ac{FastRCNN} even faster and less dependent on selective search~\cite{uijlings2013selective}, Lenc and Vedaldi (2015)~\cite{lenc2015r} used a static set of region proposals.
Ren et al. (2015)~\cite{ren2015faster} instead used the output of the existing convolutional layers plus additional convolutional layers to predict regions, resulting in a further increase in accuracy and efficiency.

Redmon et al. (2015)~\cite{redmon2015you} proposed \ac{YOLO}, which operates quicker though with less accuracy than other state-of-art approaches.
A single \ac{CNN} operates on an entire image, divided in a grid of rectangular cells, without region proposals.
Each cell outputs bounding box predictions and class probabilities;
unlike previous work, this occurs simultaneously.
Huang et al. (2015)~\cite{huang2015densebox} proposed a similar system, introducing up-sampling layers to ensure the model performs better with very small and overlapping objects.




\subsection{Cross-Depiction Detection and Matching}
Early work relating to non-photographic images focused on matching hand-drawn sketches.
Jacobs et al.\ (1995)~\cite{jacobs1995fast} used wavelet decomposition of image colour channels to allow matching between a rough colour image sketch and a more detailed colour image.
Funkhouser et al.\ (2003)~\cite{funkhouser2003search} used a distance transform of a binary line drawing, followed by fourier analysis of the distance transforms at fixed radii from the centre of the drawing, to match 2D sketches and 3D projections, with limited performance.
Hu and Collomosse (2013)~\cite{hu2013performance} used a modified version of \ac{HOG}~\cite{dalal2005histograms} to extract descriptors at interest-points in the image: for photographs, these are at Canny edges~\cite{canny1986computational} pixels; for sketches, these are sketch strokes.
Wang et al. (2015)~\cite{wang2015sketch} used a siamese \ac{CNN} configuration to match sketches and 3D model projections, optimising the \ac{CNN} to minimise the distances between sketches and 3D model projections of the same class.

Another cross-depiction matching approach, by Crowley et al. (2015)~\cite{Crowley15}, uses \ac{CNN} generated features to match faces between photos and artwork.
This relies on the success of a general face detector~\cite{parkhi2015deep}, which succeeds on artwork which is ``largely photo-realistic in nature'' but has not been verified on more abstract artwork styles such as cubism.

Other work has sought to use self-similarity to detect patterns across different depictions such as Shechtman and Irani (2007)~\cite{shechtman2007matching} and  Chatfield et al.\ (2009)~\cite{chatfield2009efficient} who used self-similarity descriptors formed by convolving small regions within in image over a larger region.
This approach is not suitable for identifying (most) objects as a whole: for example, the results show effective matching of people forming a very specific pose, not of matching people as an object class in general.

Recent work has focused on cross-depiction object classification and detection.
Wu et al.\ (2014)~\cite{wu2014learning} improved upon Felzenszwalb et al.'s \ac{DPM}~\cite{felzenszwalb2010object} to perform cross-depiction matching between photographs and ``artwork'', (including ``clip-art'', cartoons and paintings).
Instead of using root and part-based filters and a latent \ac{SVM}, the authors learnt a fully connected graph to better model object structure between depictions, using the \ac{SSVM} formulation of Cho et al.\ (2013)~\cite{cho2013learning}.
In addition, each model has separate ``attributes'' for photographs and ``artwork'': at test-time, the detector uses the maximum response from either of  ``attribute'' set, to achieve depiction invariance.
This work improved performance for detecting objects in artwork, but depended on a high performing \ac{DPM} to bootstrap the model.
Our dataset is more challenging than the one used, leading to a low accuracy using \ac{DPM} and hence this is approach is also not suitable.

Zissermann et al. (2014)~\cite{crowley2014search} evaluate the performance of \acp{CNN} learnt on \acp{photo} for classifying objects in paintings, showing strong performance in spite of the different domain.
Their evaluation excludes people as a class, as people appear frequently in their paintings without labels.
Our \textit{People-Art} dataset addresses this issue: all people are labelled and hence we provide a new benchmark.
We also believe our dataset contains more variety in terms of artwork styles and presents a more challenging problem.
Furthermore, we advance their findings: we show the performance improvement when a \ac{CNN} is fine-tuned for this task rather than simply fine-tuned on \acp{photo}.


\section{The \textit{People-Art} Dataset and its Challenges}
\label{sec:dataset}
Our \textit{People-Art} dataset\footnote{https://github.com/BathVisArtData/PeopleArt} contains images divided into 43 depiction styles.
Images from 41 of these styles came from \textit{WikiArt.org} while the \acp{photo} came from PASCAL VOC 2012~\cite{Everingham10} and the cartoons from google searches.
We labelled people since, according to our empirical observations, people are drawn or painted more often than other objects.
Consequently, this increases the total number of individual instances and thus the range of depictive styles represented.
Figure \ref{fig:oneImagePerStyle} shows one painting from each style represented in our \textit{People-Art} dataset.

The 41 depictive styles from \textit{WikiArt.org} are catagorised based on art movements.
These depiction styles cover the full range of projective and denotational styles, as defined by Willats~\cite{willats1997art}.
In addition, we propose that these styles cover many poses, a factor which Willats did not consider.

We believe that our dataset is challenging for the following reasons:
\begin{description}
\item[range of denotational styles] This is the style with which primitive marks are made (brush strokes, pencil lines, etc.)~\cite{willats1997art}.
We consider \acp{photo} to be a depictive style in its own right.
\item[range of projective style] This includes linear camera projection, orthogonal projection, inverse perspective, and in fact a range of ad-hoc projections~\cite{willats1997art}. An extreme form is shown in cubism, in which it is common for the view of a person from many different viewpoints to be drawn or painted on the 2D canvas~\cite{ginosar2014detecting}.
\item[range of poses] Though pose is handled by previous computer vision algorithms~\cite{felzenszwalb2010object}, we have observed that \ac{artwork}, in general, exhibits a wider variety of poses than \acp{photo}.
\item[overlapping, occluded and truncated people] This occurs in \ac{artwork} as in \acp{photo}, and perhaps to a greater extent.
\end{description}

\section{\ac{CNN} architecture}
\label{sec:CNNarchitecture}
\begin{figure}[b!]
  \centering
  \includegraphics[width=0.98\textwidth]{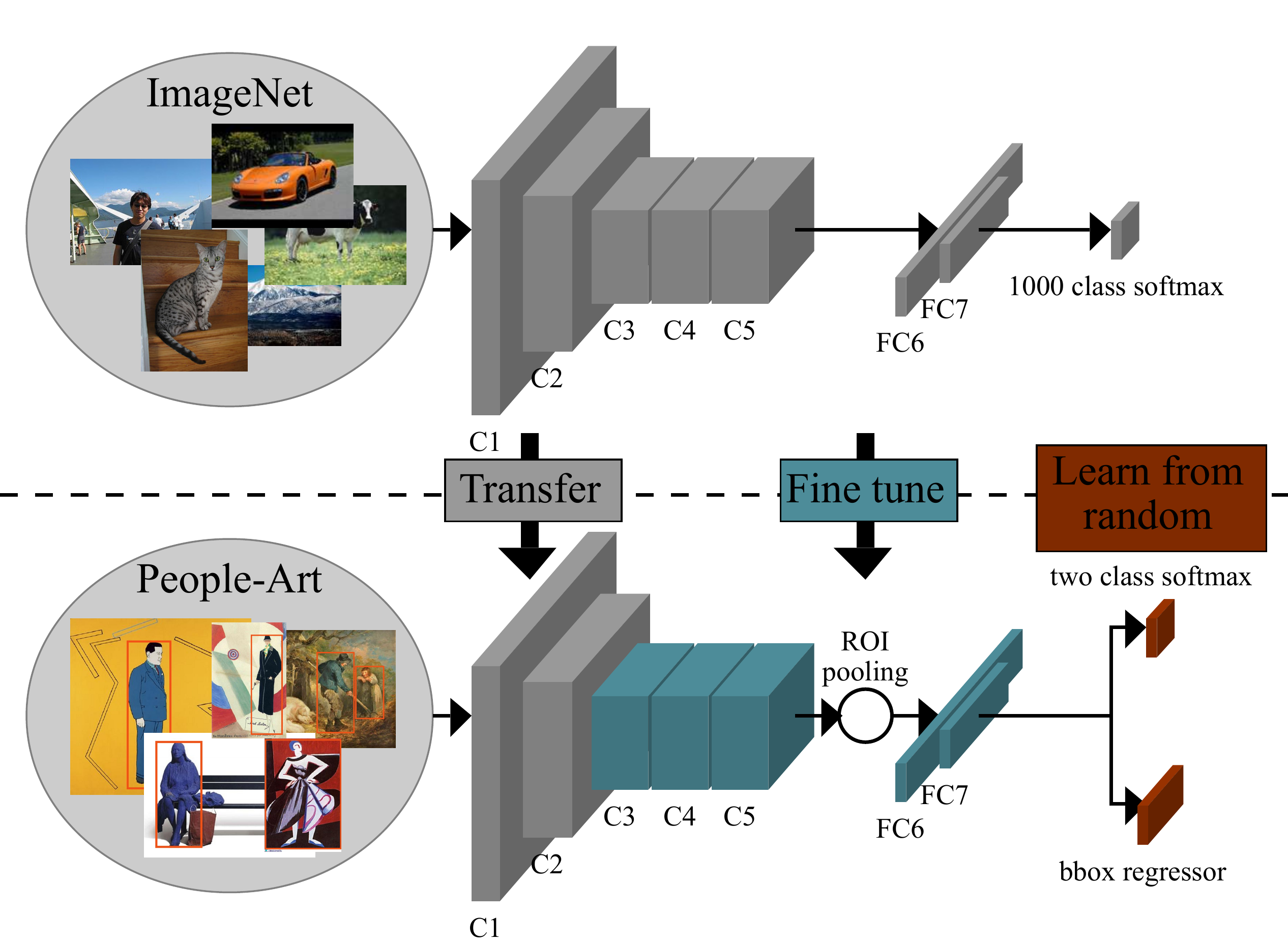}
  \caption{We use a network pre-trained on ImageNet and fine-tuned on our \textit{People-Art} dataset (training and validation sets): we fix the weights for the first F layers, selected by validation.\label{fig:networkArchitecture}}
\end{figure}

We use the same architecture as \ac{FastRCNN}~\cite{girshick2015fast}, which is built around a modified version of the Caffe library~\cite{jia2014caffe}.
The \ac{CNN} has two inputs: an image and a set of class-agnostic rectangular region proposals.
Many algorithms exist for generating region proposals; we use selective search~\cite{uijlings2013selective} with the default configuration.
\label{sec:architectureSelSearch}

The first stage of the \ac{CNN} operates on the entire image (having been resized to a fixed dimension while preserving aspect ratio).
This stage consists of convolutional layers, \acp{RELU}~\cite{krizhevsky2012imagenet,nair2010rectified}, max-pooling layers and, in some cases, local response normalisation layers~\cite{krizhevsky2012imagenet}.
The final layer is a \ac{ROI} pooling layer which is novel to \ac{FastRCNN}: as well as the input from the previous convolutional or \ac{RELU} layer, this layer receives another input, a region proposal or \ac{ROI}; the  output is a fixed-length feature vector formed by max-pooling of the convolution features.
In order to preserve information about the global structure of the \ac{ROI}, i.e.\ at a scale within an order of magnitude of the \ac{ROI} size, the max-pooling happens over a uniformly spaced rectangular grid, size
\begin{math}
H \times W
\end{math}.
As a result, the layer outputs feature vector with
\begin{math}
CHW 
\end{math}
dimensions where
\begin{math}
C    
\end{math}
is the number of channels of the previous convolutional layer.

This feature vector is the input to the second stage of the \ac{CNN}, which is fully connected.
It consists of inner product and \ac{RELU} layers, as well as dropout layers (training only) aimed at preventing overfitting~\cite{srivastava2014dropout}.
The output for each class is a score and a set of four co-ordinate which indicate the bounding box co-ordinates relative to the \ac{ROI}.
We modified the final layer to output a score and bounding box prediction for only one class: person.

We use the same approach for training as \ac{FastRCNN}, which uses \ac{SGD} with momentum~\cite{krizhevsky2012imagenet}, initialising the network with weights from the pre-trained models, in our case, trained on ImageNet~\cite{deng2009imagenet,krizhevsky2012imagenet}. 
We fix the weights of the first 
\begin{math}
  F
\end{math}
convolutional layers to those in the pre-trained model; this parameter is selected by validation.
We experiment with different criteria for the region proposals to use as training \ac{ROI}, as detailed in Section \ref{sec:ROISelection}.
Since the final inner product layers have a different size output as we only detect one class, we use random (Gaussian) initialisation.
Figure \ref{fig:networkArchitecture} shows our network architecture in detail.

We fine-tune the models (pre-trained on ImageNet) using our \textit{People-Art} dataset (training and validation sets).
We test three different models: CaffeNet, which is a reproduction of AlexNet~\cite{krizhevsky2012imagenet} with some minor changes, Oxford VGG's ``CNN M 1024'' (VGG1024) ~\cite{chatfield2014return} and Oxford VGG's ``Net D'' (VGG16)~\cite{simonyan2014very}. 
Both CaffeNet and VGG1024 have five convolutional layers and local response normalisation layers and vary slightly: in particular VGG1024 has more weights and channels.
VGG16 is much a larger network, with thirteen convolutional layers and no local response normalisation.
Except for the number of dimensions, all three networks have the same \ac{ROI} pooling layer and fully connected network structure: each CNN's fully connected network structure consists of two inner product layers, each followed by \ac{RELU} and dropout layers (training only).

\section{Experiments}
For both validation and testing, our benchmark is \acf{AP}: we calculate this using the same method as PASCAL \ac{VOC} detection task~\cite{everingham2007pascal}.
A positive detection is one whose \ac{IOU} overlap with a ground-truth bounding box is greater than 50\%;
duplicate detections are considered false.
Annotations marked as difficult are excluded.

\subsection{\ac{ROI} Selection and Layer Fixing for \ac{CNN} Fine-Tuning}
\label{sec:ROISelection}

\begin{table}[b!]
\begin{center}
\caption{Validation performance using different criteria for positive and negative \ac{ROI}: we use CNNs pre-trained on ImageNet, fine-tune on the training set and then test on the validation set; we select the best configuration for each \ac{CNN} (bold). \label{tbl:ROISelection}}
\begin{tabular}{cccccc}
  \hline
  & & \multicolumn{2}{c}{\textbf{\ac{ROI} \ac{IOU}}} \\
  \cline{3-4}
  \textbf{CNN} & \textbf{configuration} & \textbf{negative} & \textbf{positive} &
  \textbf{fixed layers (F)} & \textbf{AP} \\
 \hline
  CaffeNet & default &\begin{math} {[{0.1},{0.5})} \end{math} & \begin{math}\geq 0.5 \end{math} & 2 & 33.7\% \\
  CaffeNet & gap &\begin{math} {[{0.1},{0.4})} \end{math} & \begin{math}\geq 0.6 \end{math} & 2 & 33.5\% \\
\textbf{  CaffeNet } & \textbf{ all-neg } & \begin{math}\mathbf{ {[{0.0},{0.5})} }\end{math}  &  \begin{math}\mathbf{\geq 0.5 }\end{math}  & \textbf{ 0 } & \textbf{ 42.5\%} \\
  CaffeNet & gap + all-neg &\begin{math} {[{0.0},{0.4})} \end{math} & \begin{math}\geq 0.6 \end{math} & 1 & 42.2\% \\
\hline
  VGG1024 & default &\begin{math} {[{0.1},{0.5})} \end{math} & \begin{math}\geq 0.5 \end{math} & 1 & 38.4\% \\
  VGG1024 & gap &\begin{math} {[{0.1},{0.4})} \end{math} & \begin{math}\geq 0.6 \end{math} & 3 & 35.8\% \\
\textbf{  VGG1024 } & \textbf{ all-neg } & \begin{math}\mathbf{ {[{0.0},{0.5})} }\end{math}  &  \begin{math}\mathbf{\geq 0.5 }\end{math}  & \textbf{ 1 } & \textbf{ 42.6\%} \\
  VGG1024 & gap + all-neg &\begin{math} {[{0.0},{0.4})} \end{math} & \begin{math}\geq 0.6 \end{math} & 1 & 42.0\% \\
\hline
  VGG16 & default &\begin{math} {[{0.1},{0.5})} \end{math} & \begin{math}\geq 0.5 \end{math} & 1 & 43.9\% \\
  VGG16 & gap &\begin{math} {[{0.1},{0.4})} \end{math} & \begin{math}\geq 0.6 \end{math} & 2 & 39.0\% \\
  VGG16 & all-neg &\begin{math} {[{0.0},{0.5})} \end{math} & \begin{math}\geq 0.5 \end{math} & 3 & 50.0\% \\
\textbf{  VGG16 } & \textbf{ gap + all-neg } & \begin{math}\mathbf{ {[{0.0},{0.4})} }\end{math}  &  \begin{math}\mathbf{\geq 0.6 }\end{math}  & \textbf{ 3 } & \textbf{ 50.1\%} \\
\hline
\end{tabular}
\end{center}
\end{table}

Although we used the default selective search settings to generate region proposals, we experimented with different criteria to specify which region proposals to use in training.
The default configuration of Fast-RCNN~\cite{girshick2015fast} defines positive \ac{ROI} be region proposals whose \ac{IOU} overlap with a ground-truth bounding box is at least
\begin{math}
  0.5
\end{math},
 and defines negative \ac{ROI} to be those whose overlap lies in the interval 
\begin{math}
  {[{0.1},{0.5})}
\end{math}.
The cutoff between positive and negative \ac{ROI} matches the definition of positive detection according the \ac{VOC} detection task~\cite{everingham2007pascal}.
Girshick (2015) states that the lower cut-off 
(\begin{math}
0.1
\end{math})
for negative \ac{ROI} appears to act as a heuristic to mine hard examples~\cite{girshick2015fast,felzenszwalb2010object}.

We experimented with two alternative configurations for fine tuning:
\begin{description}
  \item[gap] We discarded \ac{ROI} whose \ac{IOU} overlap with a ground-truth bounding box lies in the interval
    \begin{math}
      {[ {0.4},{0.6})}
    \end{math}: 
    we hypothesised that \ac{ROI} lying in this interval are ambiguous and hamper training performance. 
  \item[all-neg] We removed the lower bound for negative \ac{ROI}.
    We hypothesised that this would improve performance on our \textit{People-Art} dataset for two reasons:
    \begin{enumerate}
      \item This results in the inclusion of \ac{ROI} containing classes which appear similar to people, for example animals with faces.
      \item This permits the inclusion of more artwork examples, for example images without any people present.
            We hypothesised that this would make the \ac{CNN} better able to discern between features caused by the presence of people and features resulting from a particular depiction style.
    \end{enumerate}
\end{description}
We fixed all other hyper-parameters of the \ac{CNN} except for
\begin{math}
  F
\end{math},
the number of convolutional layers whose weights we fix to those learnt from ImageNet, which we select based validation performance.

Table \ref{tbl:ROISelection} shows the validation performance for the different criteria, i.e.\ from testing on the validation set after fine-tuning on the \textit{People-Art} training set.
Removing the lower bound on negative \ac{ROI} (all-neg) results in a significant increase in performance, around a 9 percentage point increase in average precision in the best performing case.
Indeed, it appears that what is \emph{not} a person is as important as what \emph{is} a person for training.
Discarding \ac{ROI} with an \ac{IOU} overlap in the interval
\begin{math}
  {[ {0.4},{0.6})}
\end{math}
yields mixed results: it was marginally beneficial in one case, and detrimental in all others.

We note that the optimal number of convolutional layers for which to fix weights to the pre-trained model,
\begin{math}
  F
\end{math}, varies across the different training configurations, even for the same \ac{CNN}.
The variation in performance could be explained by stochastic variation caused by the use of \ac{SGD}.
The performance falls rapidly for
\begin{math}
F \geq 5
\end{math};
we therefore conclude that the first three or four convolutional layers transfer well from \acp{photo} to artwork.
Fine-tuning these layers yields no significant improvement nor detriment in performance.
In this respect, we show similar results to Yosinski et al.\ (2014)~\cite{yosinski2014transferable} for our task: i.e.\ the first three or four convolutional layers are more transferable than later layers, in our case from \acp{photo} to \ac{artwork}.

For all later experiments, including the performance benchmarks, we select the configuration which maximises performance on the \textit{validation set} (bold in Table \ref{tbl:ROISelection}) and re-train (fine-tune) using the combined \textit{train and validation sets}.

\subsection{Performance Benchmarks on the People-Art Dataset}
\label{sec:peopleArtBenchmarks}
\begin{table}[b]
\begin{center}
\caption{Performance of different methods on the test set of our \textit{People-Art} dataset: the best performance is achieved using a \ac{CNN} (Fast R-CNN) fine-tuned on \textit{People-Art}\label{tbl:PerformanceBenchmarks}}
\begin{tabular}{cccc}
  \hline
  & \multicolumn{2}{c}{\textbf{datasets}} & \\
  \cline{2-3}
  \textbf{method} & \textbf{pre-train} & \textbf{fine tuning} &
  \textbf{average precision} \\
  \hline
 Fast R-CNN (CaffeNet) & ImageNet & People-Art (train+val) & 46\% \\ Fast R-CNN (VGG1024) & ImageNet & People-Art (train+val) & 51\% \\ Fast R-CNN (VGG16) & ImageNet & People-Art (train+val) & 59\% \\  \hline
 Fast R-CNN (CaffeNet) & ImageNet & VOC 2007 & 36\% \\ Fast R-CNN (VGG1024) & ImageNet & VOC 2007 & 36\% \\ Fast R-CNN (VGG16) & ImageNet & VOC 2007 & 43\% \\  \hline
 DPM~\cite{felzenszwalb2010object} & People-Art & N/A & 33\% \\
 YOLO~\cite{redmon2015you} & ImageNet & VOC 2010 & 45\% \\
  \hline

  \end{tabular}
\end{center}
\end{table}

\begin{figure}[p]
  \centering
  \includegraphics[height=4.5cm]{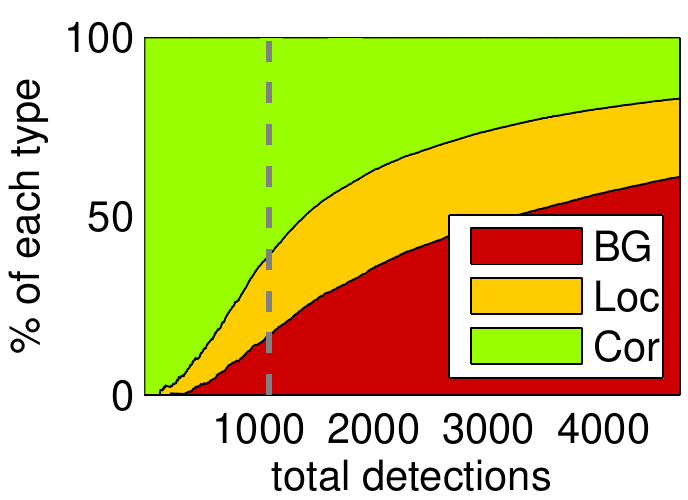} \hfill
  \includegraphics[height=4.5cm]{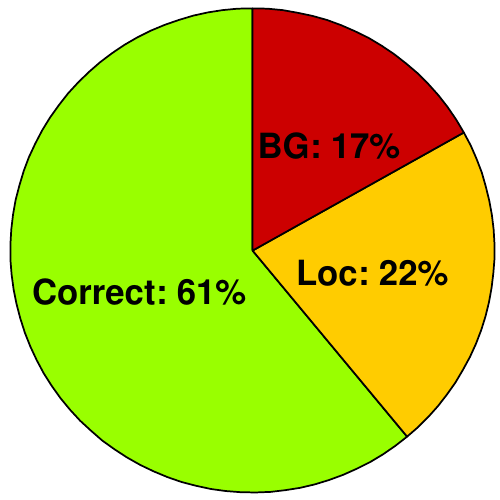}
  \caption{Left: The proportion of detections by type as the threshold decreases: either correct, a background region (BG) or poor localisation (LOC); Right: the proportion for D=1088, the actual number of people, marked as a grey dashed line on the left plot \label{fig:detectionTrend}}
  \bigskip

    \includegraphics[height=3cm]{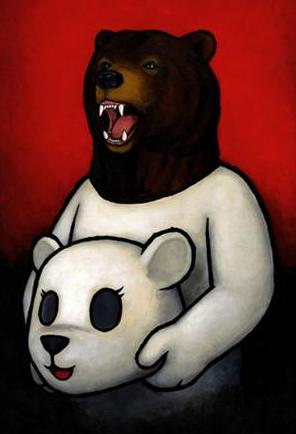} \hfill
    \includegraphics[height=3cm]{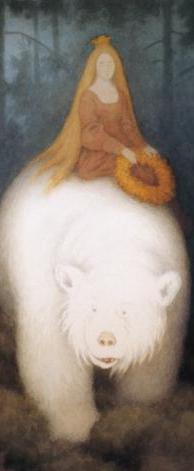} \hfill
    \includegraphics[height=3cm]{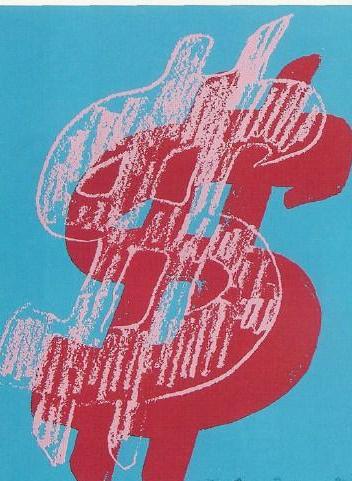} \hfill
    \includegraphics[height=3cm]{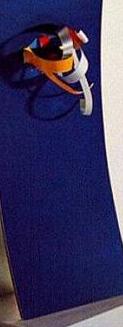} \hfill
    \includegraphics[height=3cm]{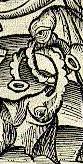} \hfill
    \includegraphics[height=3cm]{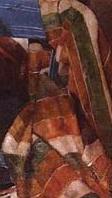} \hfill
    \includegraphics[height=3cm]{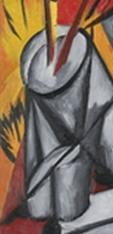} \\

    \includegraphics[height=3cm]{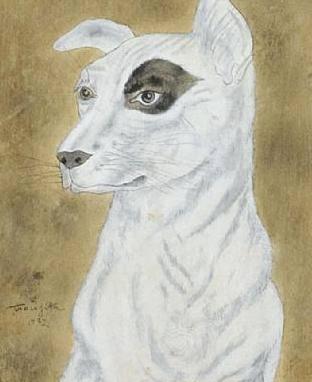} \hfill
    \includegraphics[height=3cm]{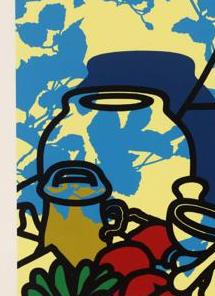} \hfill
    \includegraphics[height=3cm]{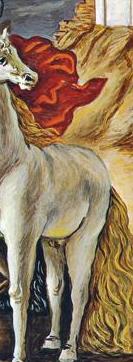} \hfill
    \includegraphics[height=3cm]{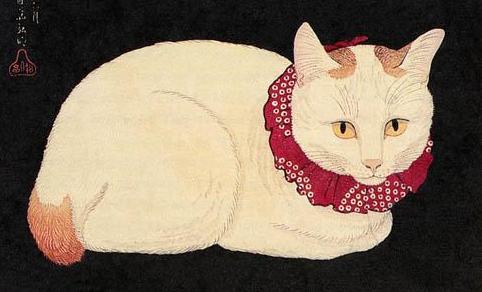}
    \caption{False positive detections on background regions from the best performing \ac{CNN}\label{fig:falsePositiveBackground}}
    \bigskip

    \includegraphics[height=2.2cm]{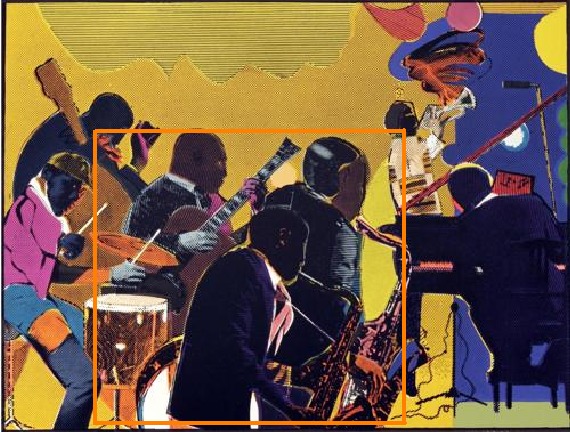} \hfill
    \includegraphics[height=2.2cm]{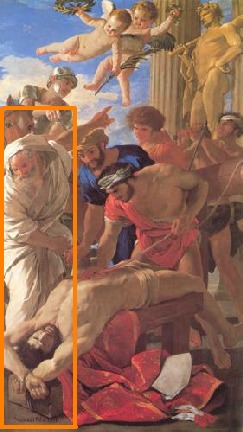} \hfill
    \includegraphics[height=2.2cm]{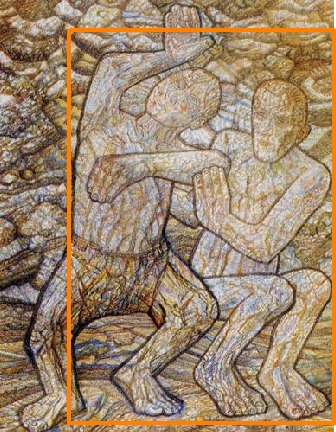} \hfill
    \includegraphics[height=2.2cm]{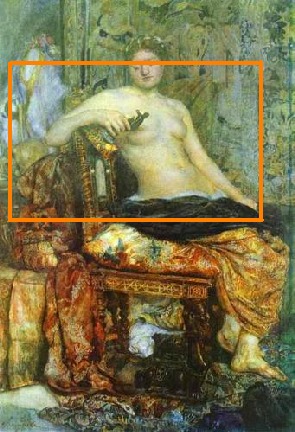} \hfill
    \includegraphics[height=2.2cm]{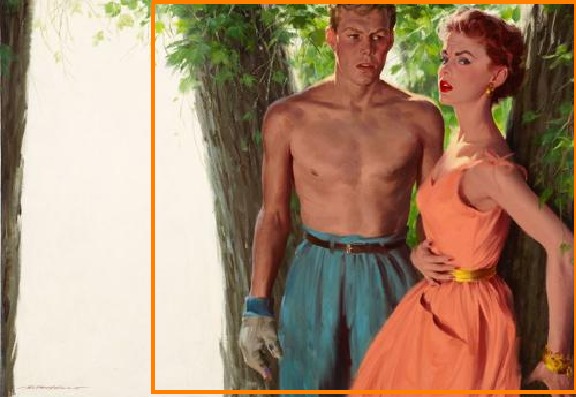} \\

    \includegraphics[height=2.2cm]{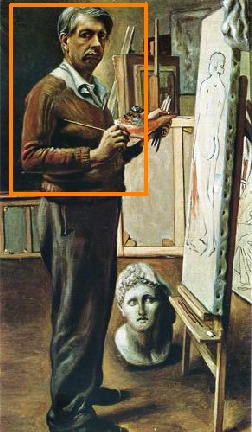} \hfill
    \includegraphics[height=2.2cm]{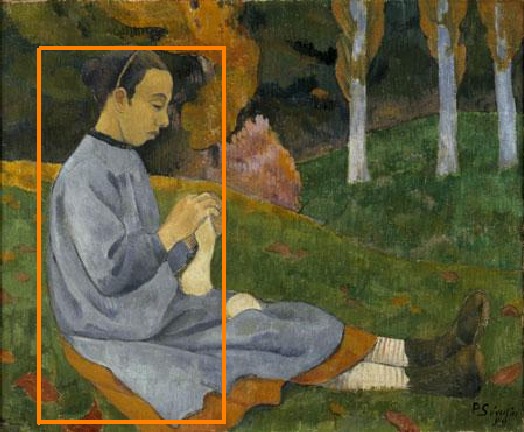} \hfill
    \includegraphics[height=2.2cm]{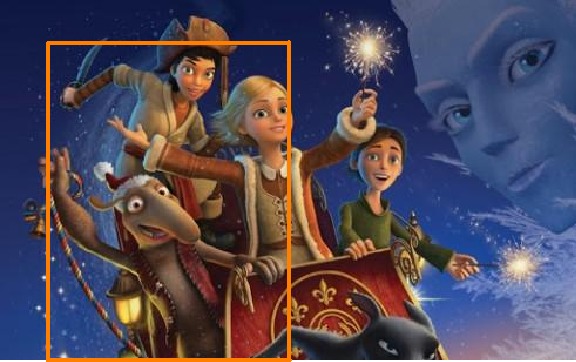} \hfill
    \includegraphics[height=2.2cm]{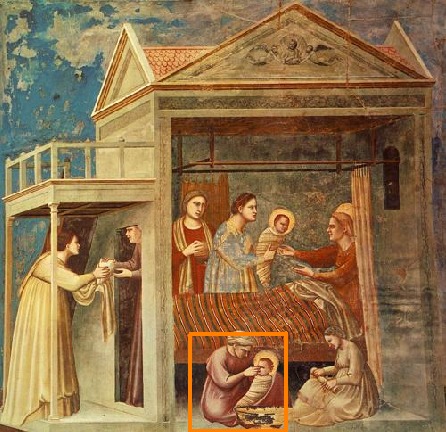} \hfill
    \includegraphics[height=2.2cm]{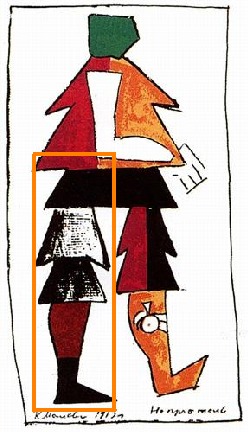}
\caption{False positive detections due to poor localisation from the best performing \ac{CNN}\label{fig:falsePositiveLocalisation}}
\end{figure}

Table \ref{tbl:PerformanceBenchmarks} shows how each \ac{CNN} model and other methods perform on the \textit{People-Art} \textit{test set}.
The best performing \ac{CNN}, VGG16, scores 58\% \ac{AP}, an improvement of 13 percentage points on the best previous result 45\%~\cite{redmon2015you}.
The results demonstrate the benefits of fine-tuning the \ac{CNN} (on the \textit{training and validation sets} of \textit{People-Art}) for the task.
We also conclude that training and fine-tuning a \ac{CNN} on \acp{photo} yields a model which overfits to photographic images.

As noted in Section \ref{sec:architectureSelSearch}, \ac{FastRCNN} (unlike \ac{YOLO}) relies on an external algorithm, here selective search~\cite{uijlings2013selective}, to generate region proposals.
We used the default settings, which are tuned to \acp{photo}.
Selective Search achieves a recall rate of 98\% on the \textit{People-Art} \textit{test set}.
As such, this does not appear to be a limiting factor for the performance.

We attempted to fine-tune \ac{YOLO}~\cite{redmon2015you} on \textit{People-Art}.
The default configuration results in an exploding gradient, perhaps due to the sparsity of regions containing objects (only people in this case) compared to other datasets.
We expect that a brute-force search over the parameters or heuristic may solve this problem in future work.

\subsection{Detection Performance on People-Art}
We used the tools of Hoiem et al. (2012)~\cite{hoiem2012diagnosing} to analyse the detection performance of the best performing \ac{CNN}.
Since we only have a single class (person), detections have three types based on their \ac{IOU} with a ground truth labelling:
\begin{description}
\item[Cor] correct i.e.\
\begin{math}
  IoU \geq 0.5
\end{math}
\item[Loc] false positive caused by poor localisation, 
\begin{math}
  0.1 \leq IoU < 0.5
\end{math}
\item[BG] a background region,
\begin{math}
  IoU < 0.1
\end{math}
\end{description}

Figure \ref{fig:detectionTrend} shows the detection trend: the proportion of detection types as the number of detections increases, i.e.\ from reducing the threshold.
At higher thresholds, the majority of incorrect detections are caused by poor localisation; at lower thresholds, background regions dominate.
In total, there are 1088 people labelled in the test set, and that are not labelled difficult.
The graph in Figure \ref{fig:detectionTrend} shows a grey dashed line corresponding to this number detections and Figure \ref{fig:detectionTrend} shows a separate pie chart for this threshold.
This threshold corresponding to this number of detections is significant: with perfect detection, there would be no false positives or false negatives.
This shows that poor localisation is the bigger cause of false positives, though only slightly more so than background regions.

Figure \ref{fig:falsePositiveBackground} shows false positives caused by background regions.
Some are caused by mammals which is understandable given these, like people, have faces and bodies.
Others detections have less clear causes.
Figure \ref{fig:falsePositiveLocalisation} show the false positives caused by poor localisation.
In some of the cases, the poor localisation is caused by the presence of more than one person, which leads to the bounding box covering multiple people.
In other cases, the bounding box does not cover the full extent of the person, i.e.\ it misses limbs or the lower torso.
We believe that this shows the extent to which the range of poses makes detecting people in \ac{artwork} a challenging problem.

\subsection{Performance Benchmarks on the Picasso Dataset}
\label{sec:PicassoBenchmarks}
\begin{table}
\begin{center}
\caption{Performance of different methods on the \textit{Picasso} dataset\label{tbl:PicassoPerformanceBenchmarks}}
\begin{tabular}{cccc}
    \hline
    \textbf{method} & \textbf{training}& \textbf{fine tuning} &
    \textbf{average precision} \\
    \hline
 Fast R-CNN (CaffeNet) & ImageNet & People-Art & 45\% \\ Fast R-CNN (VGG1024) & ImageNet & People-Art & 44\% \\ Fast R-CNN (VGG16) & ImageNet & People-Art & 44\% \\  \hline
 Fast R-CNN (CaffeNet) & ImageNet & VOC 2007 & 29\% \\ Fast R-CNN (VGG1024) & ImageNet & VOC 2007 & 37\% \\ Fast R-CNN (VGG16) & ImageNet & VOC 2007 & 33\% \\  \hline
 DPM~\cite{felzenszwalb2010object} & VOC 2007 & N/A & 38\% \\
 YOLO~\cite{redmon2015you} & ImageNet & VOC 2012 & 53\% \\
  \hline

  \end{tabular}
\end{center}
\end{table}

In addition to the results on \textit{People-Art}, we show results on the \textit{Picasso Dataset}~\cite{ginosar2014detecting}.
The dataset contains a set of Picasso paintings and labellings for people which are based on the median of the labellings given by multiple human participants.
Table \ref{tbl:PicassoPerformanceBenchmarks} shows how each \ac{CNN} and other methods perform.
As before, each \ac{CNN} performed better if it was fine-tuned on \textit{People-Art} rather than \textit{VOC 2007}; moreover, \ac{DPM} performs better than \acp{CNN} fine-tuned on \textit{VOC 2007} but worse than those fine-tuned on \textit{People-Art}.
This confirms our earlier findings: \acp{CNN} fine-tuned on \acp{photo} overfit to \ac{photo}.
In addition, we show that our fine-tuning results in a model which is not just better for \textit{People-Art} but a dataset containing \ac{artwork} which we did not train on.

Interestingly, the best performing \ac{CNN} is the smallest (CaffeNet), suggesting that the \acp{CNN} may still be overfitting to less abstract \ac{artwork}.
Furthermore, the best performing method is \ac{YOLO} despite being fine-tuned on \acp{photo} (\textit{VOC 2012}).
Selective Search achieved a recall rate of 99\% on the \textit{Picasso Dataset}, so this is unlikely to be the reason that \ac{FastRCNN} performs worse than \ac{YOLO}.
We therefore believe that \ac{YOLO}'s design is more robust to abstract forms of art.

\subsection{The Importance of Global Structure}
\label{sec:structure}
\begin{figure}[b]
  \hfill
  \includegraphics[height=3cm]{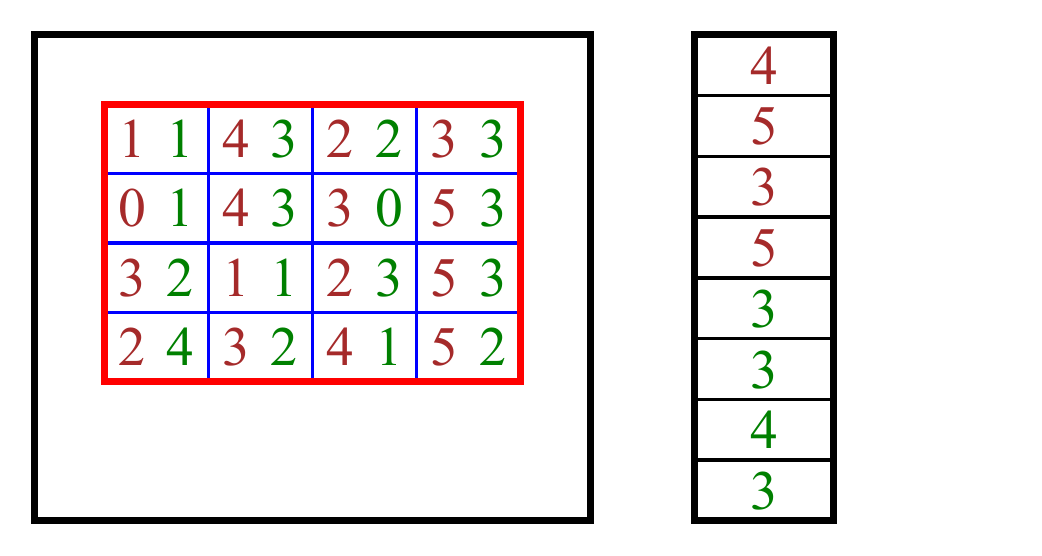} \hfill
  \includegraphics[height=3cm]{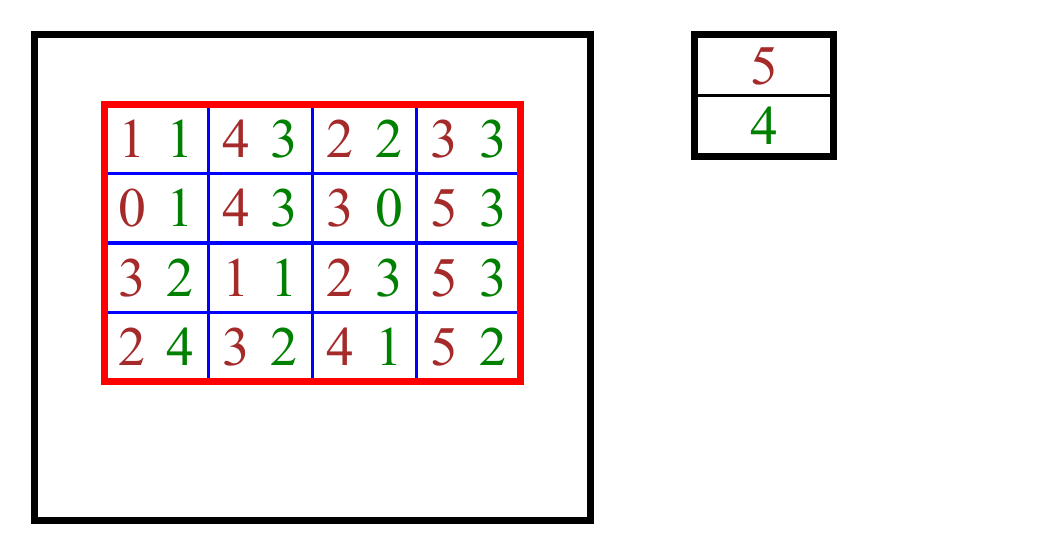}
  \hfill
  \caption{Two pooling layers and their resulting feature vectors from a two channel input; Left: An ROI pooling layer (red grid) takes the maximum for each channel in each cell of an ROI (blue grid) resulting in an 8 dimensional vector; Right: A global max-pooling layer simply takes the maximum yielding a 2 dimensional vector\label{fig:ROIPool}}
\end{figure}

\begin{table}[t]
  \begin{center}
    \caption{Replacing the \ac{ROI} pooling layer (default) with a single cell max-pooling layer yields a performance drop greater than not fine tuning \textit{People-Art} \label{tbl:PerformanceStructure}}
\begin{tabular}{c|ccc}
  \hline 

  \textbf{Fine-Tuning} & \multicolumn{2}{c}{People-Art} & VOC 2007  \\ 

  \hline 

  \textbf{ROI Pooling} & default & single cell & default  \\ 
  \hline
  CaffeNet & 46\% & 34\% & 36\% \\ 
  VGG1024 & 51\% & 35\% & 36\% \\ 
  VGG16 & 59\% & 40\% & 43\% \\ 
  \hline
\end{tabular}
  \end{center}
\end{table}

Earlier work~\cite{wu2014learning,xiao2008structure,xiao2011learning} suggests that structure is invariant across depictive styles, and therefore useful for cross-depiction detection.
As described in Section \ref{sec:CNNarchitecture}, \ac{FastRCNN} includes an \ac{ROI} pooling layer, which carries out max-pooling over
\begin{math}
H \times W
\end{math}
uniformly spaced rectangular grid.
Therefore, the \ac{ROI} pooling layer captures the global structure of the person, while earlier convolutional layers only pick up the local structure.

To test whether the \emph{global structure} is useful for detecting and localising people in \ac{artwork}, we replaced the \ac{ROI} pooling layer replaced with a single cell max-pooling layer.
This is equivalent to setting
\begin{math}
  W=1
\end{math}
and
\begin{math}
  H=1
\end{math}
for the \ac{ROI} pooling layer (see Figure \ref{fig:ROIPool}).
This is similar to ``bag of visual word'' algorithms: with
\begin{math}
W=H=1
\end{math}, the fully connected layers have no information about the location the previous layer's output.
We fine-tuned as before.

Table \ref{tbl:PerformanceStructure} shows the results.
In all cases, replacing the default \ac{ROI} pooling layer with a single cell max-pooling layer results in worse performance.
On top of this, the performance is worse than when fine-tuned on \textit{VOC 2007} with the default configuration.
This supports the claim of earlier work, that structure is invariant across depictive styles.


\section{Conclusion}
We have demonstrated state-of-the-art cross-depiction detection performance on our challenge dataset, \textit{People-Art}, by fine-tuning a \ac{CNN} for this task.
In doing so, we have shown that a \ac{CNN} trained on photograph alone overfits to \acp{photo}, while fine-turning on \ac{artwork} allows the \ac{CNN} to better generalise to other styles of \ac{artwork}.
We have also made other observations, including the importance of negative exemplars from \ac{artwork}.

The performance on our \textit{People-Art} dataset, though the best so far, is still less than 60\% \ac{AP}.
We have demonstrated that the \ac{CNN} often detects other mammals instead of people or makes other spurious detections and often fails to localise people correctly.
We propose further work to address these issues.

In addition, the dataset only covers a subset of possible images containing people.
Our dataset does not include African, Babylonian, Chinese or Egyptian art, the Bayeux Tapestry, stained glass windows, photos of sculptures and all kinds of other possibilities.
Therefore, we are only beginning to examine the cross-depiction problem, which provides a huge scope for further research.

\section*{Acknowledgements}
This research was funded in part by EPSRC grant reference EP/K015966/1.
This research made use of the Balena High Performance Computing Service at the University of Bath.




\bibliographystyle{splncs}
\bibliography{paper}

\begin{thebibliography}{10}

\bibitem{hall2015cross}
Hall, P., Cai, H., Wu, Q., Corradi, T.:
\newblock Cross-depiction problem: Recognition and synthesis of photographs and
  artwork.
\newblock Computational Visual Media \textbf{1}(2) (2015)  91--103

\bibitem{deng2009imagenet}
Deng, J., Dong, W., Socher, R., Li, L.J., Li, K., Fei-Fei, L.:
\newblock Imagenet: A large-scale hierarchical image database.
\newblock In: Computer Vision and Pattern Recognition, 2009. CVPR 2009. IEEE
  Conference on, IEEE (2009)  248--255

\bibitem{wu2014learning}
Wu, Q., Cai, H., Hall, P.:
\newblock Learning graphs to model visual objects across different depictive
  styles.
\newblock In: Computer Vision--ECCV 2014.
\newblock Springer (2014)  313--328

\bibitem{girshick2015fast}
Girshick, R.:
\newblock Fast r-cnn.
\newblock In: Proceedings of the IEEE International Conference on Computer
  Vision. (2015)  1440--1448

\bibitem{ginosar2014detecting}
Ginosar, S., Haas, D., Brown, T., Malik, J.:
\newblock Detecting people in cubist art.
\newblock In: Computer Vision-ECCV 2014 Workshops, Springer (2014)  101--116

\bibitem{yosinski2014transferable}
Yosinski, J., Clune, J., Bengio, Y., Lipson, H.:
\newblock How transferable are features in deep neural networks?
\newblock In: Advances in neural information processing systems. (2014)
  3320--3328

\bibitem{fukushima1980neocognitron}
Fukushima, K.:
\newblock Neocognitron: A self-organizing neural network model for a mechanism
  of pattern recognition unaffected by shift in position.
\newblock Biological cybernetics \textbf{36}(4) (1980)  193--202

\bibitem{giebel1971feature}
Giebel, H.:
\newblock Feature extraction and recognition of handwritten characters by
  homogeneous layers.
\newblock In: Zeichenerkennung durch biologische und technische Systeme/Pattern
  Recognition in Biological and Technical Systems.
\newblock Springer (1971)  162--169

\bibitem{lecun1989backpropagation}
LeCun, Y., Boser, B., Denker, J.S., Henderson, D., Howard, R.E., Hubbard, W.,
  Jackel, L.D.:
\newblock Backpropagation applied to handwritten zip code recognition.
\newblock Neural computation \textbf{1}(4) (1989)  541--551

\bibitem{krizhevsky2012imagenet}
Krizhevsky, A., Sutskever, I., Hinton, G.E.:
\newblock Imagenet classification with deep convolutional neural networks.
\newblock In: Advances in neural information processing systems. (2012)
  1097--1105

\bibitem{matan1992reading}
Matan, O., Baird, H.S., Bromley, J., Burges, C.J., Denker, J.S., Jackel, L.D.,
  Le~Cun, Y., Pednault, E.P., Satterfield, W.D., Stenard, C.E.,  et~al.:
\newblock Reading handwritten digits: A zip code recognition system.
\newblock Computer \textbf{25}(7) (1992)  59--63

\bibitem{nowlan1995convolutional}
Nowlan, S.J., Platt, J.C.:
\newblock A convolutional neural network hand tracker.
\newblock Advances in Neural Information Processing Systems (1995)  901--908

\bibitem{rowley1998neural}
Rowley, H.A., Baluja, S., Kanade, T.:
\newblock Neural network-based face detection.
\newblock Pattern Analysis and Machine Intelligence, IEEE Transactions on
  \textbf{20}(1) (1998)  23--38

\bibitem{sermanet2013pedestrian}
Sermanet, P., Kavukcuoglu, K., Chintala, S., LeCun, Y.:
\newblock Pedestrian detection with unsupervised multi-stage feature learning.
\newblock In: Proceedings of the IEEE Conference on Computer Vision and Pattern
  Recognition. (2013)  3626--3633

\bibitem{dalal2005histograms}
Dalal, N., Triggs, B.:
\newblock Histograms of oriented gradients for human detection.
\newblock In: Computer Vision and Pattern Recognition, 2005. CVPR 2005. IEEE
  Computer Society Conference on. Volume~1., IEEE (2005)  886--893

\bibitem{felzenszwalb2010object}
Felzenszwalb, P.F., Girshick, R.B., McAllester, D., Ramanan, D.:
\newblock Object detection with discriminatively trained part-based models.
\newblock Pattern Analysis and Machine Intelligence, IEEE Transactions on
  \textbf{32}(9) (2010)  1627--1645

\bibitem{sermanet2014overfeat}
Sermanet, P., Eigen, D., Zhang, X., Mathieu, M., Fergus, R., LeCun, Y.:
\newblock Overfeat: Integrated recognition, localization and detection using
  convolutional networks.
\newblock In: ICLR. (2014)

\bibitem{vaillant1994original}
Vaillant, R., Monrocq, C., Le~Cun, Y.:
\newblock Original approach for the localisation of objects in images.
\newblock IEE Proceedings-Vision, Image and Signal Processing \textbf{141}(4)
  (1994)  245--250

\bibitem{szegedy2013deep}
Szegedy, C., Toshev, A., Erhan, D.:
\newblock Deep neural networks for object detection.
\newblock In: Advances in Neural Information Processing Systems. (2013)
  2553--2561

\bibitem{schulz2011object}
Schulz, H., Behnke, S.:
\newblock Object-class segmentation using deep convolutional neural networks.
\newblock In: Proceedings of the DAGM Workshop on New Challenges in Neural
  Computation, Citeseer (2011)  58--61

\bibitem{girshick2014rich}
Girshick, R., Donahue, J., Darrell, T., Malik, J.:
\newblock Rich feature hierarchies for accurate object detection and semantic
  segmentation.
\newblock In: Proceedings of the IEEE conference on computer vision and pattern
  recognition. (2014)  580--587

\bibitem{uijlings2013selective}
Uijlings, J.R., van~de Sande, K.E., Gevers, T., Smeulders, A.W.:
\newblock Selective search for object recognition.
\newblock International journal of computer vision \textbf{104}(2) (2013)
  154--171

\bibitem{he2015spatial}
He, K., Zhang, X., Ren, S., Sun, J.:
\newblock Spatial pyramid pooling in deep convolutional networks for visual
  recognition.
\newblock Pattern Analysis and Machine Intelligence, IEEE Transactions on
  \textbf{37}(9) (2015)  1904--1916

\bibitem{grauman2005pyramid}
Grauman, K., Darrell, T.:
\newblock The pyramid match kernel: Discriminative classification with sets of
  image features.
\newblock In: Computer Vision, 2005. ICCV 2005. Tenth IEEE International
  Conference on. Volume~2., IEEE (2005)  1458--1465

\bibitem{lazebnik2006beyond}
Lazebnik, S., Schmid, C., Ponce, J.:
\newblock Beyond bags of features: Spatial pyramid matching for recognizing
  natural scene categories.
\newblock In: Computer Vision and Pattern Recognition, 2006 IEEE Computer
  Society Conference on. Volume~2., IEEE (2006)  2169--2178

\bibitem{lenc2015r}
Lenc, K., Vedaldi, A.:
\newblock R-cnn minus r.
\newblock arXiv preprint arXiv:1506.06981 (2015)

\bibitem{ren2015faster}
Ren, S., He, K., Girshick, R., Sun, J.:
\newblock Faster r-cnn: Towards real-time object detection with region proposal
  networks.
\newblock In: Advances in Neural Information Processing Systems. (2015)  91--99

\bibitem{redmon2015you}
Redmon, J., Divvala, S., Girshick, R., Farhadi, A.:
\newblock You only look once: Unified, real-time object detection.
\newblock arXiv preprint arXiv:1506.02640 (2015)

\bibitem{huang2015densebox}
Huang, L., Yang, Y., Deng, Y., Yu, Y.:
\newblock Densebox: Unifying landmark localization with end to end object
  detection.
\newblock arXiv preprint arXiv:1509.04874 (2015)

\bibitem{jacobs1995fast}
Jacobs, C.E., Finkelstein, A., Salesin, D.H.:
\newblock Fast multiresolution image querying.
\newblock In: Proceedings of the 22nd annual conference on Computer graphics
  and interactive techniques, ACM (1995)  277--286

\bibitem{funkhouser2003search}
Funkhouser, T., Min, P., Kazhdan, M., Chen, J., Halderman, A., Dobkin, D.,
  Jacobs, D.:
\newblock A search engine for 3d models.
\newblock ACM Transactions on Graphics (TOG) \textbf{22}(1) (2003)  83--105

\bibitem{hu2013performance}
Hu, R., Collomosse, J.:
\newblock A performance evaluation of gradient field hog descriptor for sketch
  based image retrieval.
\newblock Computer Vision and Image Understanding \textbf{117}(7) (2013)
  790--806

\bibitem{canny1986computational}
Canny, J.:
\newblock A computational approach to edge detection.
\newblock Pattern Analysis and Machine Intelligence, IEEE Transactions on (6)
  (1986)  679--698

\bibitem{wang2015sketch}
Wang, F., Kang, L., Li, Y.:
\newblock Sketch-based 3d shape retrieval using convolutional neural networks.
\newblock arXiv preprint arXiv:1504.03504 (2015)

\bibitem{Crowley15}
Crowley, E.J., Parkhi, O.M., Zisserman, A.:
\newblock Face painting: querying art with photos.
\newblock In: British Machine Vision Conference. (2015)

\bibitem{parkhi2015deep}
Parkhi, O.M., Vedaldi, A., Zisserman, A.:
\newblock Deep face recognition.
\newblock In: British Machine Vision Conference. Volume~1. (2015) ~6

\bibitem{shechtman2007matching}
Shechtman, E., Irani, M.:
\newblock Matching local self-similarities across images and videos.
\newblock In: Computer Vision and Pattern Recognition, 2007. CVPR'07. IEEE
  Conference on, IEEE (2007)  1--8

\bibitem{chatfield2009efficient}
Chatfield, K., Philbin, J., Zisserman, A.:
\newblock Efficient retrieval of deformable shape classes using local
  self-similarities.
\newblock In: Computer Vision Workshops (ICCV Workshops), 2009 IEEE 12th
  International Conference on, IEEE (2009)  264--271

\bibitem{cho2013learning}
Cho, M., Alahari, K., Ponce, J.:
\newblock Learning graphs to match.
\newblock In: Proceedings of the IEEE International Conference on Computer
  Vision. (2013)  25--32

\bibitem{crowley2014search}
Crowley, E.J., Zisserman, A.:
\newblock In search of art.
\newblock In: Workshop at the European Conference on Computer Vision, Springer
  (2014)  54--70

\bibitem{Everingham10}
Everingham, M., Van~Gool, L., Williams, C.K.I., Winn, J., Zisserman, A.:
\newblock The pascal visual object classes (voc) challenge.
\newblock International Journal of Computer Vision \textbf{88}(2) (June 2010)
  303--338

\bibitem{willats1997art}
Willats, J.:
\newblock Art and representation: New principles in the analysis of pictures.
\newblock Princeton University Press (1997)

\bibitem{jia2014caffe}
Jia, Y., Shelhamer, E., Donahue, J., Karayev, S., Long, J., Girshick, R.,
  Guadarrama, S., Darrell, T.:
\newblock Caffe: Convolutional architecture for fast feature embedding.
\newblock arXiv preprint arXiv:1408.5093 (2014)

\bibitem{nair2010rectified}
Nair, V., Hinton, G.E.:
\newblock Rectified linear units improve restricted boltzmann machines.
\newblock In: Proceedings ofs the 27th International Conference on Machine
  Learning (ICML-10). (2010)  807--814

\bibitem{srivastava2014dropout}
Srivastava, N., Hinton, G., Krizhevsky, A., Sutskever, I., Salakhutdinov, R.:
\newblock Dropout: A simple way to prevent neural networks from overfitting.
\newblock The Journal of Machine Learning Research \textbf{15}(1) (2014)
  1929--1958

\bibitem{chatfield2014return}
Chatfield, K., Simonyan, K., Vedaldi, A., Zisserman, A.:
\newblock Return of the devil in the details: Delving deep into convolutional
  nets.
\newblock arXiv preprint arXiv:1405.3531 (2014)

\bibitem{simonyan2014very}
Simonyan, K., Zisserman, A.:
\newblock Very deep convolutional networks for large-scale image recognition.
\newblock arXiv preprint arXiv:1409.1556 (2014)

\bibitem{everingham2007pascal}
Everingham, M., Winn, J.:
\newblock The pascal visual object classes challenge 2007 (voc2007) development
  kit.
\newblock University of Leeds, Tech. Rep (2007)

\bibitem{hoiem2012diagnosing}
Hoiem, D., Chodpathumwan, Y., Dai, Q.:
\newblock Diagnosing error in object detectors.
\newblock In: European conference on computer vision, Springer (2012)  340--353

\bibitem{xiao2008structure}
Xiao, B., Song, Y.Z., Balika, A., Hall, P.M.:
\newblock Structure is a visual class invariant.
\newblock In: Joint IAPR International Workshops on Statistical Techniques in
  Pattern Recognition (SPR) and Structural and Syntactic Pattern Recognition
  (SSPR), Springer (2008)  329--338

\bibitem{xiao2011learning}
Xiao, B., Yi-Zhe, S., Hall, P.:
\newblock Learning invariant structure for object identification by using graph
  methods.
\newblock Computer Vision and Image Understanding \textbf{115}(7) (2011)
  1023--1031

\end{thebibliography}

\end{document}